# Detection Transformer for Teeth Detection, Segmentation, and Numbering in Oral Rare Diseases: Focus on Data Augmentation and Inpainting Techniques

Kadi Hocine
Research and Innovation Department
Capgemini Engineering
Illkirch-Graffenstaden, France
hocine.a.kadi@capgemini.com

Sourget Théo
1) Department of Computer Science
Université Rouen Normandie
Rouen, France
2) IT University of Copenhagen
Copenhagen, Denmark
theo.sourget@univ-rouen.fr

Kawczynski Marzena
Reference Center for Rare Oral and Dental Diseases
Hôpitaux Universitaires de Strasbourg
Strasbourg, France
marzena.kawczynski@chru-strasbourg.fr

Bendjama Sara
Reference Center for Rare Oral and Dental Diseases
Hôpitaux Universitaires de Strasbourg
Strasbourg, France
sara.bendjama@gmail.com

Grollemund Bruno
Reference Center for Rare Oral and Dental Diseases
Hôpitaux Universitaires de Strasbourg
Strasbourg, France
bruno.grollemund@gmail.com

Bloch-Zupan Agnès
1) Université de Strasbourg, Faculté de Chirurgie Dentaire, Strasbourg, France
2) Hôpitaux Universitaires de Strasbourg, Pôle de Médecine et Chirurgie Bucco-dentaires, Centre de référence des maladies rares orales et dentaires, CRMR-O-Rares, Filière Santé Maladies rares TETE COU, European Reference Network ERN CRANIO, Strasbourg, France
3) Université de Strasbourg, CNRS-UMR7104, INSERM U1258, Institut de Génétique et de Biologie Moléculaire et Cellulaire (IGBMC), Illkirch, France
agnes.bloch-zupan@unistra.fr

*Abstract*— In this work, we focused on deep learning image processing in the context of oral rare diseases, which pose challenges due to limited data availability. A crucial step involves teeth detection, segmentation and numbering in panoramic radiographs. To this end, we used a dataset consisting of 156 panoramic radiographs from individuals with rare oral diseases and labeled by experts. We trained the Detection Transformer (DETR) neural network for teeth detection, segmentation, and numbering the 52 teeth classes. In addition, we used data augmentation techniques, including geometric transformations. Finally, we generated new panoramic images using inpainting techniques with stable diffusion, by removing teeth from a panoramic radiograph and integrating teeth into it. The results showed a mAP exceeding 0,69 for DETR without data augmentation. The mAP was improved to 0,82 when data augmentation techniques are used. Furthermore, we observed promising performances when using new panoramic radiographs generated with inpainting technique, with mAP of 0,76.

## I. INTRODUCTION

Rare diseases are characterized by their low prevalence and limited understanding, present a significant challenge for healthcare practitioners and researchers alike [1]. The scarcity of data and the complex nature of these conditions hinder effective diagnosis and treatment strategies, needing innovative approaches to enhance medical outcomes. In parallel, the advent of deep learning techniques has opened up exciting possibilities for various medical applications, including the segmentation of dental panoramic radiographs, which plays a crucial role in dentistry. In this context, the use of deep learning methods has the potential to revolutionize how dental professionals analyse and diagnose dental conditions by enhancing the precision and efficiency of dental image analysis. To automatically detect, recognize and number the teeth in panoramic radiographs, different segmentation models were used [2], like Faster-RCNN [3], Mask-RCNN [4] and PANet [5] which obtained good detection accuracies, but they also showed limitations in deciduous teeth numbering [6].

Transformers-based neural networks have brought about a significant transformation in several domains of artificial intelligence, such as natural language understanding, computer vision and medical images processing [7]. For teeth semantic segmentation, SWin-Unet [8] a transformer-based neural network, has been used and showed superior segmentation performance. Currently, no paper has evaluated the usage of transformers to solve teeth instance segmentation and numbering [2]. On the other hand, Detection Transformer (DETR) neural network [9] or models derived from DETR have been applied to several medical tasks such as cell segmentation [10], lymph node detection [11], and polyp detection [12].

In the context of small dataset, various approaches have been proposed to increase the number of dental panoramic radiographs [13,14], including geometric-based approaches and classical image processing methods. Generative neural networks offer promising opportunities in addressing complex medical image analysis tasks, such as denoising, reconstruction and segmentation, data generation, detection, or classification of medical images. Several generative models have been used for medical image generation [15,16], but none, to our knowledge, in the synthesis and generation of dental panoramic radiographs for segmentation purposes.

In this article, we aim to address two key questions. The first question investigates the performances of DETR in the context of teeth instance segmentation. The second question, propose to evaluate how generative methods utilizing inpainting can enhance the performance of segmentation models by generating new panoramic radiographs, and then compare these outcomes with those achieved through traditional data augmentation techniques.

## II. DATASET

### A. Data description

The data used in our study were provided by the Reference Center for Rare Oral and Dental Diseases, CRMR O-Rares, Hôpitaux Universitaires de Strasbourg (HUS), France and performed in collaboration with Université de Strasbourg and the Institute of Genetics and Molecular and Cellular Biology (IGBMC/CERBM). Each radiograph was annotated and numbered by experts to accurately segment and classify each apparent tooth. we used genetic testing called Next Generation Sequencing with the GenoDENT panel [17] and found issues in genes like *EDA, PAX9* and *WNT10*. We removed some unusable radiographs and kept 156. The limited amount of images associated with the variability due to the medical condition and the large number of classes (52 types of teeth) which are in different stages of development represent the main difficulties of the task as the objective is not only to detect the teeth but classify them using the ISO 3950 numbering system (see Fig. 1).

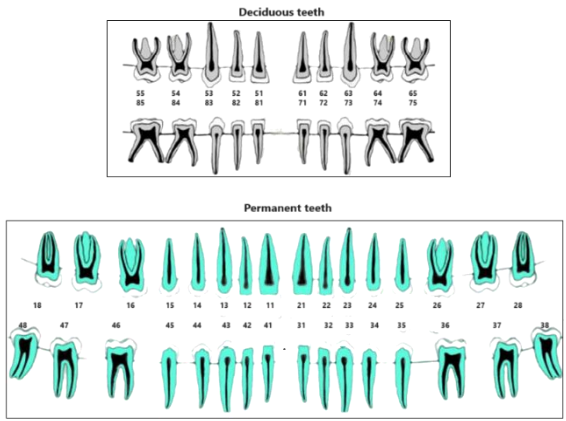

Fig. 1: ISO 3950 teeth numbering system with deciduous teeth on the top and definitive teeth on the down.

### B. Data preparation

The dataset was split into training, validation, and test dataset. An image has multiple teeth, however, for reasons of age or dental agenesis, some teeth are missing. We can see the dataset as being multi-label. Therefore, iterative stratification for multi-label data from [18] was used to split the dataset.

Fig. 2 shows the datasets distribution with 72 images in the train set, 48 in the validation and 36 in the test. The distribution inside the validation set is not completely satisfactory especially for deciduous teeth but the test set was prioritised to have a final comparison as robust as possible.

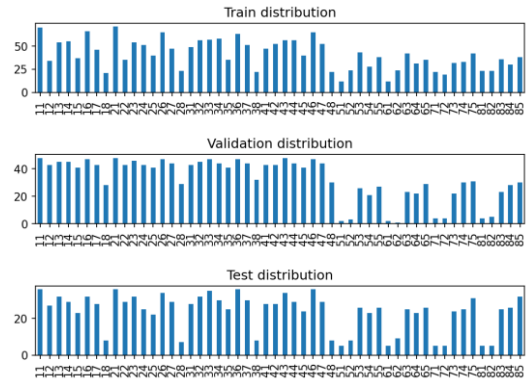

Fig. 2: Teeth distribution in split dataset

It is interesting to note that due to the age and dental agenesis of patients with oral rare diseases, certain teeth are less represented.

Pre-processing steps have been applied to each radiograph. First the histograms are equalized to enhance the contrast, then a crop around oral cavity is applied. Finally, vertical padding and resizing are used to having the same size of 1024x1024.

## III. DATA AUGMENTATION

One of the main difficulties of our problem is the limited amount of data (72 in the training set). Moreover, some studies have shown the good impact of data augmentation for teeth numbering as in [13,14]. Therefore, this part of the study will focus on 2 different approaches to augment the dataset's size.

### A. Geometrics transformations

To begin, data augmentation using some transformations on our selected dataset have been used both on the training and the validation set. 2 sets of transformations have been tested:

• Configuration 1: Rotation (range +/- 10°), Noise, Contrast

• Configuration 2: Configuration 1 + Translation (range +/- 0,05 on both axis)

For each configuration, 2 strategies have been used to generate the images. As shown in Fig. 7 there are less images with deciduous teeth so while with the first strategy five images are generated regardless of the class within the original image, with the second strategy, 5 images are generated if the image contains a deciduous tooth and 2 otherwise. At most, this adds 360 images in the training set and 240 in the validation set. We hypothesise that it would help the classification as the distribution between the classes is more balanced. An example of obtained image can be seen Fig. 3.

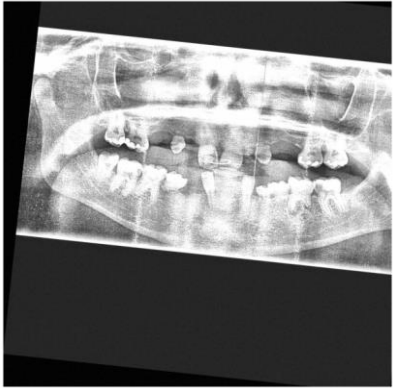

Fig. 3: Image obtained with configuration 2 of data augmentation.

*B. New panoramic radiographs with Inpainting*

Using generative methods to synthesize new images requires large amounts of data to train the models which is why we opted for a new approach. In our dataset, we collected the information about the teeth present or absent for a patient, but we have not an associated/usable panoramic radiograph. To generate new radiograph with that information we need:

1. Panoramic radiographs without any tooth: "empty" panoramic
2. Teeth to put in the empty panoramic radiograph

To generate empty radiographs, an inpainting model have been applied. We have tried two already trained models: Stable Diffusion and Stable Diffusion 2 [19]. As shown in Fig. 4, results with Stable Diffusion 1 were not satisfying while on the other hand, a majority of generated panoramic radiographs from Stable Diffusion 2 were usable to produce new radiographs.

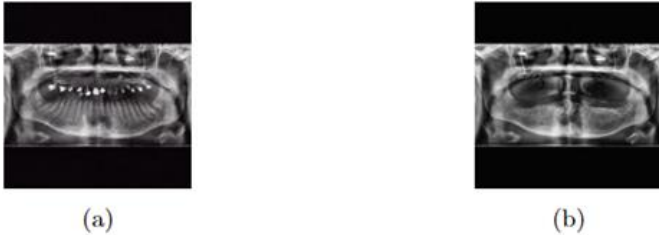

Fig. 4: Comparison between Stable Diffusion 1 (a) and Stable Diffusion 2(b) to generate empty panoramic with inpainting.

For the teeth to add in the panoramic radiographs, we use the teeth taken individually from the panoramic radiographs in the training set, using contours labels.

Having both the empty radiographs and the individual teeth we can generate new panoramic radiographs using the information about which teeth are present and the age of the patient. The procedure is the following:

1. Choose an empty radiograph of the same age, if possible, otherwise choose one randomly.
2. For each present tooth, select one from the same age, if possible, otherwise choose one randomly.
3. Add the tooth in the empty panoramic.
4. Generate the ground truth in the COCO format

Fig. 5 is the visual representation of the process and Fig. 6 shows an example of generated radiograph using this technique. We can see that the generation produced some artifacts probably due to the compression. Moreover, there are some overlaps on certain teeth, this situation is due to the fact that we used the same coordinates as in the original panoramic of the tooth but as the teeth may come from different sources the coordinates may overlap.

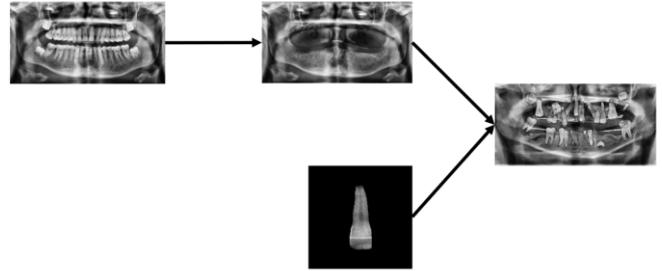

Fig. 5 Scheme of panoramic radiograph generation process

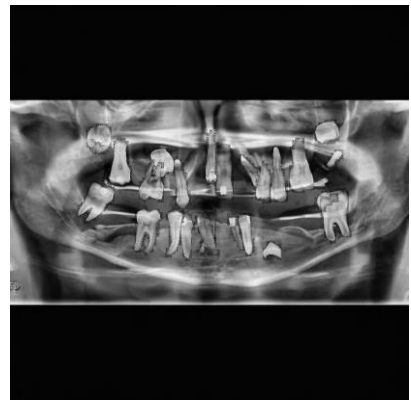

Fig. 6: Example of panoramic radiographs generated with the usage of existing teeth.

In the end, 298 new images were generated and added to the training set. As shown in Fig. 7, the number of deciduous teeth in these panoramic radiographs is limited so it's unlikely to help on the detection and numbering of these teeth, but we hypothesis that the increased number of combinations of teeth will improve the performances on the definitive ones.

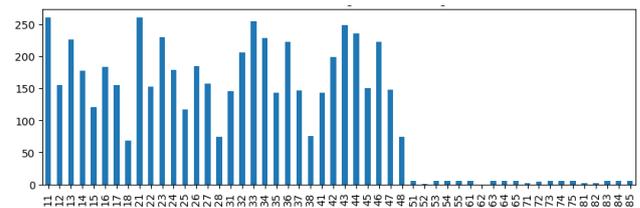

Fig. 7: Distribution of teeth in generated panoramic radiographs.

IV. METHODS AND EXPERIMENTATIONS

*A. DETR*

Presented in 2020, DETR differs from previous instance segmentation models such as Mask-RCNN in its loss function and the usage of an encoder-decoder transformer, which simplifies the overall detection process and relies less on prior knowledge.

**Operations of the model**

First, a CNN is applied to the base image to generate a lower resolution feature map. Next, a 1x1 convolution is used to further reduce the size of the feature map. The feature map is then flattened so that it is no longer a sequence of matrices but a sequence of 1d vectors, and the positional encoding of transformers is added. Then, with the encoder-decoder, a better representation can be obtained thanks to attention. Finally, the N vectors generated by the decoder are passed into a 3-layer neural network that predicts the coordinates of the boxes and, a single linear layer for the object classes and a segmentation head to produce segmentation maps.

So as stated by the authors, the model works quite simply, using only known operations, which means that any usual Deep Learning library can be used to implement or use it.

**Loss function**

In addition to using a transform encoder-decoder, the authors of DETR have defined a special loss function called Hungarian loss which enables the model to learn to predict the right number of bounding boxes. Therefore, it does not have to use an algorithm such as "Non-Maximum Suppression" to eliminate unnecessary boxes, as in R-CNN.

The loss function relies on a matching algorithm between the prediction of the model and the ground truth. This algorithm computing the best matching between the ground truth, and the predictions takes in consideration bounding box for which an object is present but also added "No-object" bounding box so there are N ground truth object, N being the number of outputs of the model for every image. This technique allows the model to be invariant to the order in the ground-truth and to make the model learn the right number of bounding boxes. With the mapping, the loss can be computed, it is composed by two or three terms depending on the task:

- the first term, a negative log likelihood, is to evaluate the classification of the bounding box.
- the second term, a combination of GIoU and L1 loss, is to evaluate the coordinates of the bounding box.
- the third term, used when training the segmentation part, is a Dice loss to evaluate the segmentation masks.

*B. Experimentations*

There are two main strategies for training the model. The first one is to start by only training the detection part of DETR using the first loss and then freeze the weights to only train the segmentation branch using the loss with Dice. The second one is to train both parts at the same time using directly the second loss. From our experimentations, we've chosen to use the second option and train the model in two steps as the training in one step resulted in longer training time.

Using pretrained weights on the coco dataset, the model is trained without the segmentation head for 300 epochs with a learning rate of $1e^{-5}$ for the CNN backbone, $1e^{-4}$ for the rest of the network and a batch size of 2. Then, the weights are frozen except for the segmentation head which is added and the training with a learning rate of $1e^{-3}$ and the same batch size is done for 300 epochs when using the base dataset and 200 epochs when using a dataset with augmentation.

We used a DGX server to train the models. This server consists of 40 CPUs, 252 GB of RAM, and four NVIDIA Tesla V100 DGXS 32GB graphics cards, each with 32 GB of VRAM. The server runs on a Linux version 4.15.0-50-generic. Only one of the four GPUs is used during training.

*C. Metrics*

The metric used to compare models in object detection and instance segmentation is the mean average precision (mAP). The mAP use both the precision and the recall to represent the quality of the model to detect objects, those 2 metrics need to know if a prediction is accurate or not. In the context of object detection and instance segmentation, a prediction is considered as good if it fulfils the 2 following criteria:
1. The IoU between the predicted bounding-box/mask and the ground-truth is above a selected threshold (0.5 in our case)
2. The predicted class is correct

*D. Post-processing*

As displayed by the multiple results obtained, the main point of improvement for this problem is on the classification part. We note that the errors are of 3 main categories.
1. Detection problem: true negatives and false positives
2. Deciduous teeth misclassification
3. Symmetric teeth misclassification

Post-processing operations can be applied to reduce errors from categories 1 and 2. For symmetric teeth, we can divide the oral cavity in 4 parts as show in Fig. 8, then depending on the coordinates of the bounding box's center we can correct the class. For example, if a tooth in the first section is classified as class 24 it will be corrected to 14.

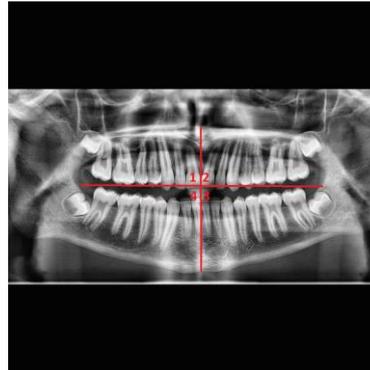

Fig. 8: Separation of the oral cavity in 4 parts to correct symmetric teeth misclassification.

One type of false detection is if the same class is detected multiple times. In this situation only the prediction with the highest confidence score will be kept and the other ones removed.

V. RESULTS AND DISCUSSION

Table I shows the results of the various configurations on the test dataset after applying post-processing. We obtained a mAP of 0,68 using DETR trained on training dataset without data augmentation. All configuration using geometric transformations or inpainting has a significant improvement on mAP, detection, and numbering. The configuration with only rotation as a geometrical transformation and additional panoramics with deciduous teeth (config 1 bis) shown the best improvement, with an increase of 20%. There is however no significant difference between the different configurations using geometric transformations and the hypothesis about the prioritisation of deciduous teeth was not totally verified.

Finally, the configuration using inpainting show promising results as it is close to the configurations with geometrical transformations. From the table, it is visible that the main weakness is the classification, we hypothesis that the defaults in the generation mentioned in III.B can be the cause of this problem. This early-stage result can however confirm that this technic can be an asset in the augmentation of the dataset size especially in this context of available data for rare diseases without associated panoramic radiographs.

TABLE I. RESULTS OF DIFFERENT MODELS ON TEST DATASET AFTER POST-PROCESSING

| Model | Training time | mAP (IoU:0.5) | Detection Accuracy | Classification Accuracy |
|---|---|---|---|---|
| DETR (without augmentation) | 11 hours | 68,6% | 85,4% | 77% |
| DETR augment config 1 | 1 day and 23 hours | 80,5% | 96,2% | 88,1% |
| **DETR augment config 1 bis** | 2 days 8 hours | **82%** | **96,7%** | **91,9%** |
| DETR augment config 2 | 1 day and 23 hours | 79,5% | 95,8% | 86,5% |
| DETR augment config 2 bis | 2 days 8 hours | 75,8% | 95,1% | 85,8% |
| DETR generated | 19 hours | 75,8% | 95,2% | 84,8% |

An example of result produced by the model (config 1 bis) can be seen in Fig. 9. This example highlights the accuracy of the model with most teeth being correctly detected and classified. However, it also shows few detection errors for small and indistinct teeth such as teeth 18, 38 and 48 in the panoramic radiograph.

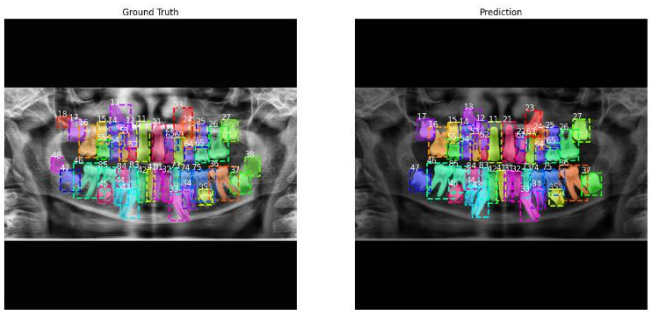

Fig. 9: Example of results with DETR augment config 1 bis.

The differences between the confusion matrix before and after the post-processing in Fig. 10 and Fig. 11, obtained with best model (config 1 bis), show the effectiveness of the method especially on symmetric misclassification. It is however visible that the method struggles with teeth close to the frontier of decision such as teeth 11, 21, 31 and 41.

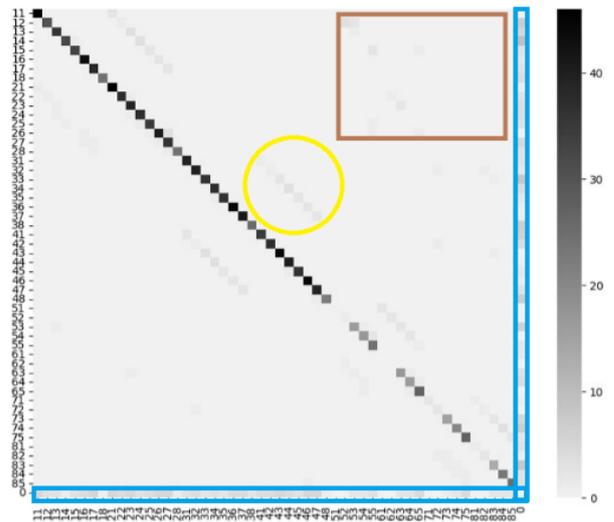

Fig. 10 Confusion matrix before post-processing with Deciduous errors (brown area), detection errors (blue area), symmetric errors (yellow area).

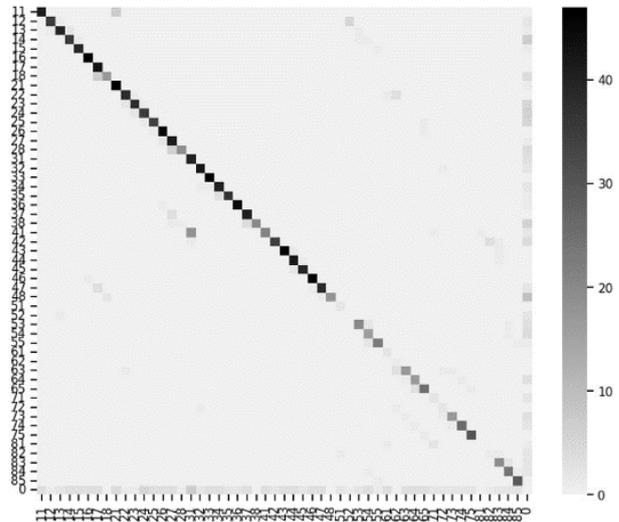

Fig. 11 Confusion matrix after post-processing.

For future works, it seems important to improve the generation of new panoramic radiographs using inpainting to confirm the usability of this technic. Moreover, we used already existing teeth to fill an empty radiograph, but it could be interesting to try with the generation of new teeth using generative models.

ACKNOWLEDGMENT

We acknowledge patients and families for their trust, participation and continuous support in our projects. We also thank G. Icre, C. Englender, C Lecomte, the CRIH team (HUS), Biovalley France, SATT Conectus (M. David), for their help throughout this project.